\pdfoutput=1
\documentclass[final]{cvpr}

\usepackage{times}
\usepackage{epsfig}
\usepackage{graphicx}
\usepackage{amsmath}
\usepackage{amssymb}

\usepackage{caption}  
\usepackage{amsfonts}
\usepackage{subcaption}
\usepackage{mathtools}
\usepackage{booktabs}
\usepackage{algorithm}
\usepackage{algorithmic}
\usepackage{mathrsfs} 
\usepackage{soul}
\usepackage{color}
\usepackage{multirow}
\usepackage{xcolor,colortbl}
\usepackage{soul} 
\usepackage{lipsum}
\usepackage{newtxtext,newtxmath}
\usepackage{comment}

\definecolor{vyellow}{rgb}{0.7490,0.5647,0}
\definecolor{vyelloworig}{rgb}{1,0.7529,0}
\definecolor{vmagenta}{rgb}{0.4392,0.1882,0.6274}
\definecolor{vmagenta3}{rgb}{0.7254,0.6352,0.7960}
\definecolor{vmagentaorig}{rgb}{0.9215,0.8039,1}
\definecolor{vblue}{rgb}{0.3686,0.7921,0.8431}
\definecolor{vblue2}{rgb}{0.062,0.1529,0.7254}
\definecolor{vred}{rgb}{0.7529,0,0}
\definecolor{darkGreen}{rgb}{0,0.6,0}
\definecolor{mistyrose}{rgb}{1.0, 0.89, 0.88}
\newcommand{\nnet}{\emph{RefVAE}\@\xspace}

\usepackage[pagebackref=true,breaklinks=true,colorlinks,bookmarks=false]{hyperref}

\def\cvprPaperID{99} 
\def\confYear{CVPR 2021}

\begin{document}

\title{Variational AutoEncoder for Reference based Image Super-Resolution}

\author{Zhi-Song Liu$^1$\\
$^1$Caritas Institute of Higher Education\\
{\tt\small zsliu@cihe.edu.hk}
\and
Wan-Chi Siu$^{1,2}$~~~~~~~~~~~~~~~~~~~~~~~~~~Li-Wen Wang $^{2}$\\
$^{1,2}$The Hong Kong Polytechnic University\\
{\tt\small enwcsiu@polyu.edu.hk, liwen.wang@connect.polyu.hk}
}

\maketitle
\thispagestyle{empty}
\pagestyle{empty}

\begin{abstract}
   In this paper, we propose a novel reference based image super-resolution approach via Variational AutoEncoder (\nnet). Existing state-of-the-art methods mainly focus on single image super-resolution which cannot perform well on large upsampling factors, e.g., 8$\times$. We propose a reference based image super-resolution, for which any arbitrary image can act as a reference for super-resolution. Even using random map or low-resolution image itself, the proposed RefVAE can transfer the knowledge from the reference to the super-resolved images. Depending upon different references, the proposed method can generate different versions of super-resolved images from a hidden super-resolution space. Besides using different datasets for some standard evaluations with PSNR and SSIM, we also took part in the NTIRE2021 SR Space challenge~\cite{ntire2021} and have provided results of the randomness evaluation of our approach. Compared to other state-of-the-art methods, our approach achieves higher diverse scores.
\end{abstract}

\section{Introduction}
\label{sec:intro}
\pdfoutput=1
Image Super-Resolution (SR) is a fundamental problem in image processing. Given a low-resolution (LR) image, the objective is to upsample the LR image by $\alpha\times$ to obtain the SR image. In other words, each LR pixel is used for predicting $\alpha\times\alpha$ HR pixels. In real applications, super-resolution is useful for images/videos display, storage, broadcasting and transmission. In particular, it can provide better visual experience for watching movies and playing video games.~\cite{nvidia,amd}

\begin{figure}[t]
\setlength{\abovecaptionskip}{0pt} 
\begin{center}
\includegraphics[width=\columnwidth]{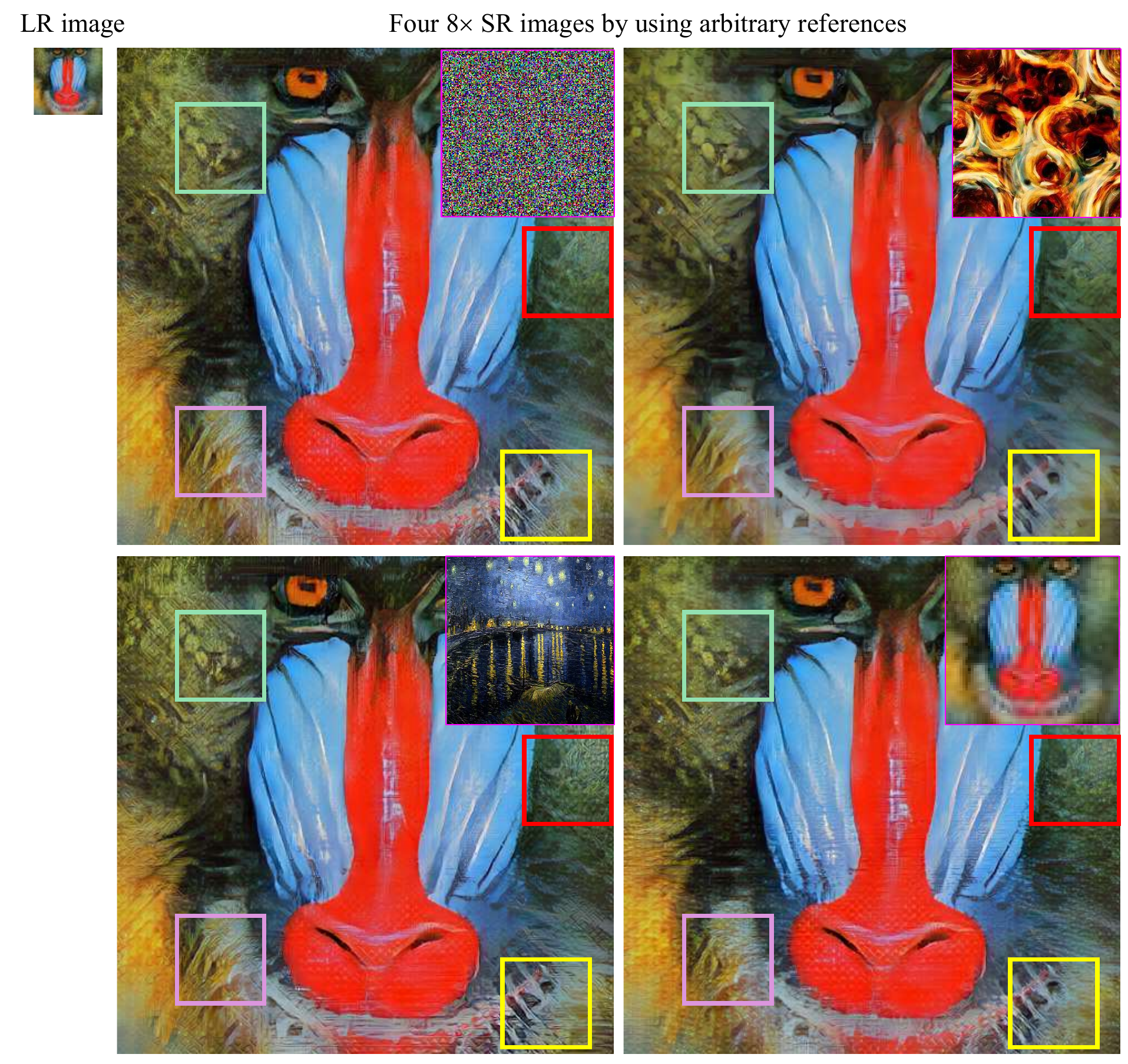}
\end{center}
  \caption{\small{\textbf{Visualization of 8$\times$ SR using our proposed \nnet}. The visualization of the SR images given different references (right upper corner of the SR images), like random noise, random images or LR image itself.}
  }
\label{fig:teaser}
\vskip -0.1in
\end{figure} 

Most works for image super-resolution are limited as they focus on single image super-resolution by optimizing the mean squared errors (MSE) between HR and SR pixels. Due to the ill-posed nature of super-resolution, using MSE usually leads to the blurry effect on edges~\cite{tradeoff,SRGAN}. One solution is to use Generative Adversarial Network to implicitly align the distribution between SR and HR images. Despite recent development on GAN based image SR~\cite{SRGAN,ESRGAN,esrgan+}, they usually focus on smaller upsampling factors, like 4$\times$~\cite{ESRGAN}. For larger upsampling factors, like 16$\times$ SR, researchers either use 4$\times$ SR model twice, or use simple interpolation to upsample the LR image by 4$\times$ first and then use SR model for another 4$\times$ upsampling~\cite{EDSR}. Our concern is that the single image super-resolution has reached its limit on large upsampling factors even with most advanced deep learning techniques. Hence, we propose a reference based image super-resolution (\nnet), which can search similar patterns from the reference to guide the super-resolution. It is an online searching but fast processing because of the propsoed Conditional Variational Inference (CVAE). It is not an uncommon procedure in image SR using online approach to search similar patches from external~\cite{refsr1} or internal~\cite{refsr4} datasets for SR. The difference is that we compress various reference images into a compact hidden space by CVAE. Instead of learning complex regression models~\cite{refsr5}, we propose to use CVAE to learn explicit distribution from reference images, then we sample the patterns from the distribution as a condition or prior to super-resolve LR image. Different from reference based image super-resolution,hence it is fast for real-time applications. Our proposed method is also different from single image SR. Once the training is done, we can use different ``references'' to generate different SR candidates, hence, we can expand the SR space to allow users to pick one of the SR candidates for applications. As shown in Figure~\ref{fig:teaser}, the ``reference'' (right-hand upper sub-images) can be any images, even a random Gaussian map or the LR image itself. Our approach can generate a SR image with good visual quality but with different perceptual details around the edges and textures (The differences are marked in boxes with different colors). 

To sum up, our contributions include:
\begin{itemize}
    \item We propose \nnet to explicitly discover image distribution using the proposed Conditional Variational AutoEncoder for image super-resolution.
	\item Instead of learning directly the data distribution, we introduce references as a condition to guide super-resolution. To transfer the features from style images, we propose to combine both pixel loss and style loss for training. (See Section )
	\item We provide experimental results on different datasets and show analysis and comparison on both single image SR and reference based image SR results. 
\end{itemize}

\section{Related work}
\label{sec:related}
\pdfoutput=1
In this section, we give a brief review on related previous works on single image super-resolution. To introduce our proposed \nnet method, we also revisit some representative works on generative learning approaches for image SR. Finally, we also introduce related works on reference based image super-resolution.

\subsection{Single image super-resolution}
Single image super-resolution uses one single LR image to produce the corresponding SR image. It is a classic topic in image processing and a lot of works have been proposed to resolve it. With the development of learning approaches, learning based image SR dominates this field. We can categorize learning based SR into two groups: conventional learning based approaches~\cite{A+,SRRMF,CRFSR} and deep learning based approaches~\cite{ABPN,RCAN,EDSR,HBPN,Kim15}. 

Let us focus on some representative deep learning based approaches. Dong et al.~\cite{SRCNN} proposed the first CNN based image SR using only three layers of convolutions to learn the mapping relations between LR and HR images. Later on, VDSR~\cite{Kim15}, LapSRN~\cite{LapSRN}, SRResNet~\cite{SRGAN}, EDSR~\cite{EDSR} and many other works~\cite{RCAN, DenseSR} use a deeper and wider (number of filters) CNN model to learn the mapping functions for super-resolution. Back projection in CNN~\cite{DBPN,HBPN} also sets a new path for image SR for which iterative residual update can minimize the feature distance for super-resolution. Recent studies~\cite{ABPN,SAN,HAN} show that attention can learn non-local features for super-resolution, especially useful for larger images. Liu et al.~\cite{ABPN} proposed an attention based back projection network for large-scale super-resolution. Dai et al.~\cite{SAN} proposed higher-order attention using channel and spatial attention for image. To fully explore the correlations in the channel dimension, Niu et al.~\cite{HAN} proposed a holistic attention network for image super-resolution. The above mentioned approaches train different convolutional neural network end-to-end to optimize mean squared errors between SR and HR pixels. The problem is that using MSE to give equal weights to different pixels can cause blurry effect. To address this problem, many researchers have proposed generative learning approaches for image SR.

\subsection{Generative learning approaches for SR}
Using generative learning approaches for image SR is a popular topic. We can consider using a single CNN model for SR as discriminative learning that does not explore the data distribution for modelling. On the other hand, generative learning approaches explicitly or implicitly study the data distribution for modelling. Generative Adversarial Network~\cite{GAN} is one of the popular implicit generative approaches. It is also used in image SR. For example, SRGAN~\cite{SRGAN} uses GAN to train 4$\times$ SR. In order to generate more photo-realistic features, VGG based feature loss is used to minimize the deep feature distances between SR and HR images. ESRGAN~\cite{ESRGAN} proposes to use relativistic GAN~\cite{relativistic} to let the discriminator predict relative realness instead of the absolute value. ESRGAN+~\cite{esrgan+} further improves ESRGAN by introducing noise for to explore stochastic variations. Vartiaonal AutoEncoder is another choice for SR.

Different from GAN, VAE explicitly explores the data distribution for modelling. Liu et al.~\cite{SRVAE} first proposed VAE for image super-resolution that achieves comparable performance as CNN based approaches. Furthermore, ~\cite{RefSR} introduces a reference based face SR that uses the conditional VAE to achieve domain-specific super-resolution. A sampling generator is proposed to constraint the choice of samples for better face reconstruction. To avoid the blurry effect~\cite{Latent_VAE} caused by the sampling process in VAE, ~\cite{dSRVAE} uses VAE for both image super-resolution and denoising by combining GAN and VAE together so that it can use adversarial loss to encourage photo-realistic feature reconstruction. Most recently, Lugmayr et al.~\cite{SRFlow} proposed to use conditional normalizing flow to model image distribution for super-resolution and achieved good performance in both quantitative and qualitative results. Normalizing flow based approaches can also be extended to image manipulation~\cite{BahatM20Explorable,BuhlerRT20DeepSEE}. However, the whole normalizing flow structure is over-complex (approx. 50,000,000 parameters) which could be difficult for training.

\subsection{Reference based Super-Resolution (RefSR)}
The development of reference based image super-resolution (RefSR) is not a surprise to researchers. Originally, it comes from the non-local filtering. The conventional learning based approaches are patch based process. In other words, the LR patches can be reconstructed by learning the mapping models from images/videos with similar contents~\cite{refsr1,refsr2,refsr3,refsr4,refsr5}. For example, k Nearest Neighbor (kNN)~\cite{refsr5} can be used for online search so it limits the big data search. Random Forests~\cite{CRFSR} is a fast algorithm that can classify training patches into different groups for diverse regression modelling. 

\begin{figure*}[t!]
\setlength{\abovecaptionskip}{0pt} 
\begin{center}
\includegraphics[width=0.9\textwidth]{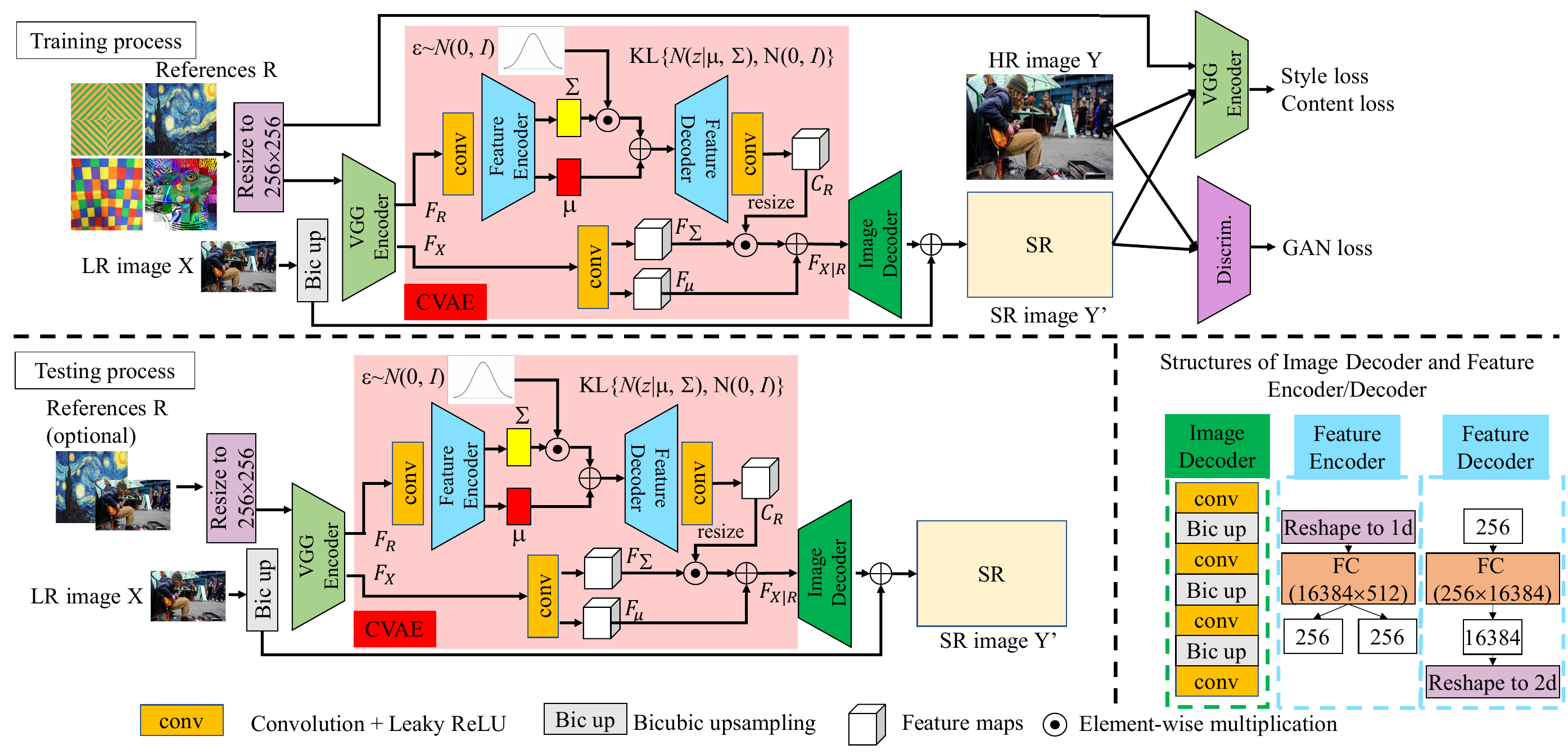}
\end{center}
  \caption{\small{\textbf{The training and testing processes of the proposed RefVAE}. It consists of 1) VGG Encoder (the green box), 2) Conditional Variational Variational AutoEncoder (CVAE) (the pink box), 3) Image Decoder (the dark green box), and Discriminator (Discrim.) for loss computation. Detailed structures of the Image Decoder and Feature Encoder/Decoder are also shown on the right bottom corner.}
  }
\label{fig:main}
\end{figure*} 

Reference based SR can also be achieved by using deep learning. ~\cite{refsr1} proposes to use an extra HR reference to guide SR. The idea is to use CNN to extract multi-scale HR features to fill the missing information for LR images. Similarly, Zheng et al. ~\cite{refsr2,refsr3} proposed a similar reference based SR approach using CNN. The difference is that it uses flow warping operation to align LR and reference features for super-resolution. However, it still requires a pre-aligned reference for SR, or the warping operation cannot well match the feature maps. Most recently, Tan et al.~\cite{refsr3} proposed a reference pool to expand the searching region for multi-reference based super-resolution. Not only CNN can be used for RefSR, Liu et al.~\cite{RefSR} proposed a face SR that uses an image from the same identity as reference. The idea is to train a conditional Variational AutoEncoder to learn conditional distribution between the reference and LR image. The limitation is that it only investigates the facial image SR and the reference image has to be the face with the same identity. It is not suitable for general image SR.

The study of RefSR also reveals the ill-posed nature of SR that more than one SR image can be formed from a LR image. Depending on different external or internal images as reference, we can transfer reference features to fill out the missing information for LR image enlargement. As described in the Learning the Super-Resolution Space Challenge~\cite{ntire2021}, we can formulate the SR problem as learning a stochastic mapping, capable of sampling from the space of plausible high-resolution images given a low-resolution image. By using an arbitrary reference image to expand data diversity, RefSR is one way to resolve SR space. The advantage is that 1) multiple predictions can be generated and compared and 2) we allow users to choose references with desired patterns for interactive SR.

\section{Method}
\label{sec:method}
\pdfoutput=1
In this section, we will give detailed introduction on the proposed Reference based image super-resolution approach via Variational AutoEncoder (\nnet). Let us formally define the image SR. Mathematically, given a LR image $\mathbf{X}\in\mathbb{R}^{\mathit{m}\times\mathit{n}\times3}$ which has been down-sampled from the corresponding HR image $\mathbf{Y}\in\mathbb{R}^{\mathit{\alpha m}\times\mathit{\alpha n}\times3}$, where ($\mathit{m}$, $\mathit{n}$) is the dimension the image and $\alpha$ is the up-sampling factor. They are related by the following degradation model,

\begin{small}
    \vskip -0.01in
	\begin{equation}
	\mathbf{X}=\mathbf{D}\otimes\mathbf{Y}+\mu \tag{1}
	\label{equation 1}
	\end{equation}
	\vskip -0.01in
\end{small}

\noindent where $\otimes$ represents the down-sampling convolution, $\mu$ is the additive white Gaussian noise and \textbf{D} denotes the down-sampling kernel. The goal of image SR is to resolve Equation~\ref{equation 1} as the Maximum A Posterior (MAP) problem as,

\begin{small}
    \vskip -0.01in
	\begin{equation}
	\mathbf{\hat{Y}}=\underset{\mathbf{Y}}{\arg\max}\, log P(\mathbf{X|Y})+log P(\mathbf{Y}) \tag{2}
	\label{equation 2}
	\end{equation}
	\vskip -0.01in
\end{small}

\noindent where $\mathbf{\hat{Y}}$ is the predicted SR image. log$P$($\mathbf{X}|\mathbf{Y}$) represents the log-likelihood of LR images given HR images and log$P$($\mathbf{Y}$) is the prior of HR images that is used for model optimization. Note that Equation~\ref{equation 2} generally describes the optimization process that multiple SR $\mathbf{\hat{Y}}$ can be found. We can rewrite Equation~\ref{equation 2} as Reference based SR,

\begin{small}
	\begin{equation} \tag*{(3)}
	\begin{matrix}
	\begin{split}
	& \mathbf{\hat{Y}_i}=\underset{\mathbf{Y}}{\arg\max}\, log P(\mathbf{X}, \mathbf{R}_i|\mathbf{Y})+log P(\mathbf{Y}) \\
	& where\ \mathbf{R}_i \in \mathbb{S}_\mathbf{R}\ and\  \mathbf{\hat{Y}_i} \in \mathbb{S}_\mathbf{Y}
	\label{equation 3}
	\end{split}
	\end{matrix} 
	\end{equation}
\end{small}

\noindent and we define the SR space as $\mathbb{S}_\mathbf{Y}$ and the reference space as $\mathbb{S}_\mathbf{R}$. The desired SR image can be randomly sampled from $\mathbb{S}_\mathbf{R}$ space given arbitrary reference image.

The complete architecture of the proposed \nnet is shown in Figure~\ref{fig:main}, which includes the training  (upper half of the figure) and testing (lower half of the figure) stages. Our proposed \nnet takes arbitrary references $\mathbf{R}_i$ and LR images $\mathbf{X}$ for training and testing. It includes three components: 1) VGG Encoder, 2) CVAE (in the pink box), and 3) Image Decoder. The VGG Encoder is used to extract feature maps from LR and reference images. CVAE is used to transfer reference features to a hidden space for conditional feature sampling. In other words, CVAE generates a candidate feature pool where random feature vectors can be sampled as conditional features. The LR feature maps are split into mean and variance maps, then they compute with the conditional features to obtain the estimated features. Finally, the Image Decoder takes the estimated features for image reconstruction.

\subsection{VGG Encoder}
The VGG Encoder follows the structure of VGG-19 by keeping all convolution layers and discarding the fully connection layers. We directly use pre-trained VGG19~\cite{VGG} to extract feature maps from references ($F_\mathbf{R}=G(\mathbf{R})$) and the bicubic upsampled LR images ($F_\mathbf{X}=G(\mathbf{X})$), where G stands for the process of VGG feature extraction. Inside the VGG Encoder, there are three maxpooling layers to downsample the input image by 8$\times$. Note that we resize arbitrary reference images to $256\times256$ before passing it through the VGG Encoder. Hence we have fixed reference feature maps as $F_\mathbf{R} \in \mathbb{R}^{32\times32\times512}$. For LR images, we initially upsample them by Bicubic interpolation to the desired size. The reason of using pre-trained VGG Encoder is that:  1) VGG was trained for general image classification that the extracted feature maps are generalized to images with different contents and 2) we want to project the LR and reference images to a same feature domain such that we can fuse their features together for super-resolution. 

\subsection{Conditional Variational AutoEncoder}
We have the Conditional Variational AutoEncoder (CVAE) that projects the reference feature maps to a latent space to learn the hidden distribution via Feature Encoder. The Feature Decoder learns to transfer the reference features as conditions $C_R$ for LR feature maps. The combination of Feature Encoder and Decoder (for detailed structures: see the right of Figure~\ref{fig:main}) forms the Variational Inference process. The idea of Variational Inference is to learn the generative model for the reference images that can be represented by a Gaussian model as $P(z|\mathbf{R})=N\sim(z;\ \mu(\mathbf{R}),\Sigma(\mathbf{R}))$, where $\mu$ and $\Sigma=diag(\sigma_1^2,...,\sigma_n^2)$ are the mean and variance of the learned Gaussian model. In other words, we learn the pixel inter-correlations and represent them as a probability model. We introduce randomness by the sampling process. As the Gaussian curve shown in Figure~\ref{fig:main}, we use a normal distribution $Q(z)=N\sim(0, I)$ to sample from the reference model as $z=\mu+\epsilon \cdot \sigma$. To ensure the learned probability model close to a normal distribution, we use KL divergence to optimize the model as,

\begin{small}
    \vskip -0.01in
	\begin{equation} \tag*{(4)}
	\begin{matrix}
	\begin{split}
	D_{KL}(P(z|\mathbf{R})& || Q(z))=E[log P(z|\mathbf{R})-Q(z)] \\
	& = \frac{1}{2}\left(-\sum_i (log\sigma^2_i+1) + \sum_i \sigma^2_i + \sum_i \mu_i^2 \right)
	\label{equation 4}
	\end{split}
	\end{matrix} 
	\end{equation}
	\vskip -0.01in
\end{small}

During the testing, we can discard the Feature Encoder as many existing VAE based image reconstruction~\cite{sanet,Johnson}. We can use vectors sampled from the normal distribution as conditional priors for super-resolution. We can also keep the Feature Encoder to allow user to define their own reference image for customized super-resolution. That is, we keep the Feature Encoder to extract specific prior distribution $P(z|\mathbf{R})=N\sim(z;\ \mu(\mathbf{R}),\Sigma(\mathbf{R}))$ as prior conditions for super-resolution. Next, to obtain the conditional feature maps $C_\mathbf{R}$, we use one convolution block to project the learned distribution back to the spacial domain (we also resize it to the same size as the LR feature map via simple interpolation).

In order to transfer the conditional features to the LR feature map, we use a convolution block to learn the mean and variance (note that the mean and variance are the spatial statistics of the feature maps, rather than the variables of the Gaussian distribution) for the LR feature maps as $F_{\mu}$ and $F_{\Sigma}$. We then have the fused features as $F_{X|R}=C_R\cdot (1 + F_{\Sigma}) + F_{\mu}$. 

\subsection{Image Decoder}
Finally, the Image Decoder learns to reconstruct the SR image from the fused feature maps  $\mathbf{\hat{Y}}$. The structure of Image Decoder (dark green boxes in Figure~\ref{fig:main}) has a similar structure as the VGG Encoder stacking three convolution layers followed by a simple bilinear interpolation.

\subsection{Training losses}
To train the proposed \nnet to generate SR results with photo-realistic visual quality, we suggest to use a discriminator to reduce the perceptual distance between SR and ground truth images. We design the same discriminator as PatchGAN~\cite{pix2pixhd} and the adversarial loss is defined as,

\begin{small}
    \vskip -0.01in
	\begin{equation} \tag*{(5)}
	L_{adv}=log[1-D(\mathbf{\hat{Y}})]
	\label{equation 5}
	\end{equation}
	\vskip -0.01in
\end{small}

The idea of using style and content losses for style transfer~\cite{sanet,Johnson} is an efficient approach to transfer the reference style to the target image while preserving the content information. It is also suitable for reference based SR. We not only want the SR image close to the ground truth, but we also want it close to the reference image in terms of style similarity. In other words, we want to ensure the reference features to be transferred to the LR images. 

\noindent \textbf{Content loss} For content loss, we extract features for the HR image \textbf{Y} and the SR $\mathbf{\hat{Y}}$ using VGG-19~\cite{VGG} (we take the feature map at \textit{relu4\_1} layer). We refer these features to as $W_\mathbf{Y}$  and $W_\mathbf{\hat{Y}}$, respectively. For pixel-wised difference, we also have the $L_1$ loss between SR and HR and their down-sampled versions. Totally, we have the content loss as,

\begin{small}
    \vskip -0.01in
	\begin{equation} \tag*{(6)}
	L_{\text{content}} = \left\Vert W_\mathbf{Y}-W_\mathbf{\mathbf{\hat{Y}}}\right\Vert^1 + \left\Vert \mathbf{Y}-\mathbf{\hat{Y}}\right\Vert^1 + \left\Vert Down(\mathbf{Y})-Down(\mathbf{\hat{Y}}) \right\Vert^1~~.
	\label{eq:Equation6}
	\end{equation}
	\vskip -0.01in
\end{small}

\noindent where \textit{Down} is the $\alpha \times$ bicubic down-sampling operator. 

\noindent \textbf{Style loss}. To measure the style similarity, we use VGG-19 to extract feature maps (\textit{relu1\_2}, \textit{relu2\_2}, \textit{relu3\_4}, \textit{relu4\_1}) for the reference image \textbf{R} and the SR image $\mathbf{\hat{Y}}$ as  $V_\mathbf{R}$ and $V_\mathbf{\hat{Y}}$. Similar to ~\cite{sanet}, we measure their mean and variance to align the SR features close to the reference features as:

\begin{small}
    \vskip -0.01in
    \begin{equation} \tag*{(7)}
	L_{\text{style}} = \sum_i \Vert mean(V^i_\mathbf{Y}-V^i_\mathbf{\hat{Y}})\Vert^1 + \Vert var(V^i_\mathbf{Y}-V^i_\mathbf{\hat{Y}})\Vert^1~~.  
	\label{eq:Equation7}
	\end{equation}
	\vskip -0.01in
\end{small}

\noindent where \textit{mean} and \textit{std} are the operations for calculating the mean and variance of the feature maps. In order to have SR images visually close to the HR image, the LPIPS loss~\cite{LPIPS} is used to measure perceptual differences as $L_{LPIPS}(\mathbf{\hat{Y}}, \mathbf{Y})$. The Total Variation loss $L_{TV}(\mathbf{\hat{Y}})$ is used to encourage smooth quality, for which we calculate the first-order horizontal and vertical pixel gradients as 

\begin{small}
\vskip -0.01in
\begin{equation} \tag*{(8)}
	L_{TV}=\sum_{i,j} \left((\mathbf{\hat{Y}}_{i,j-1}-\mathbf{\hat{Y}}_{i,j})^2 + (\mathbf{\hat{Y}}_{i-1,j}-\mathbf{\hat{Y}}_{i,j})^2 \right)^\beta~~.  
	\label{eq:Equation8}
	\end{equation}
	\vskip -0.01in
\end{small}

\begin{figure}[t]
\setlength{\abovecaptionskip}{0pt} 
\begin{center}
\includegraphics[width=0.9\columnwidth]{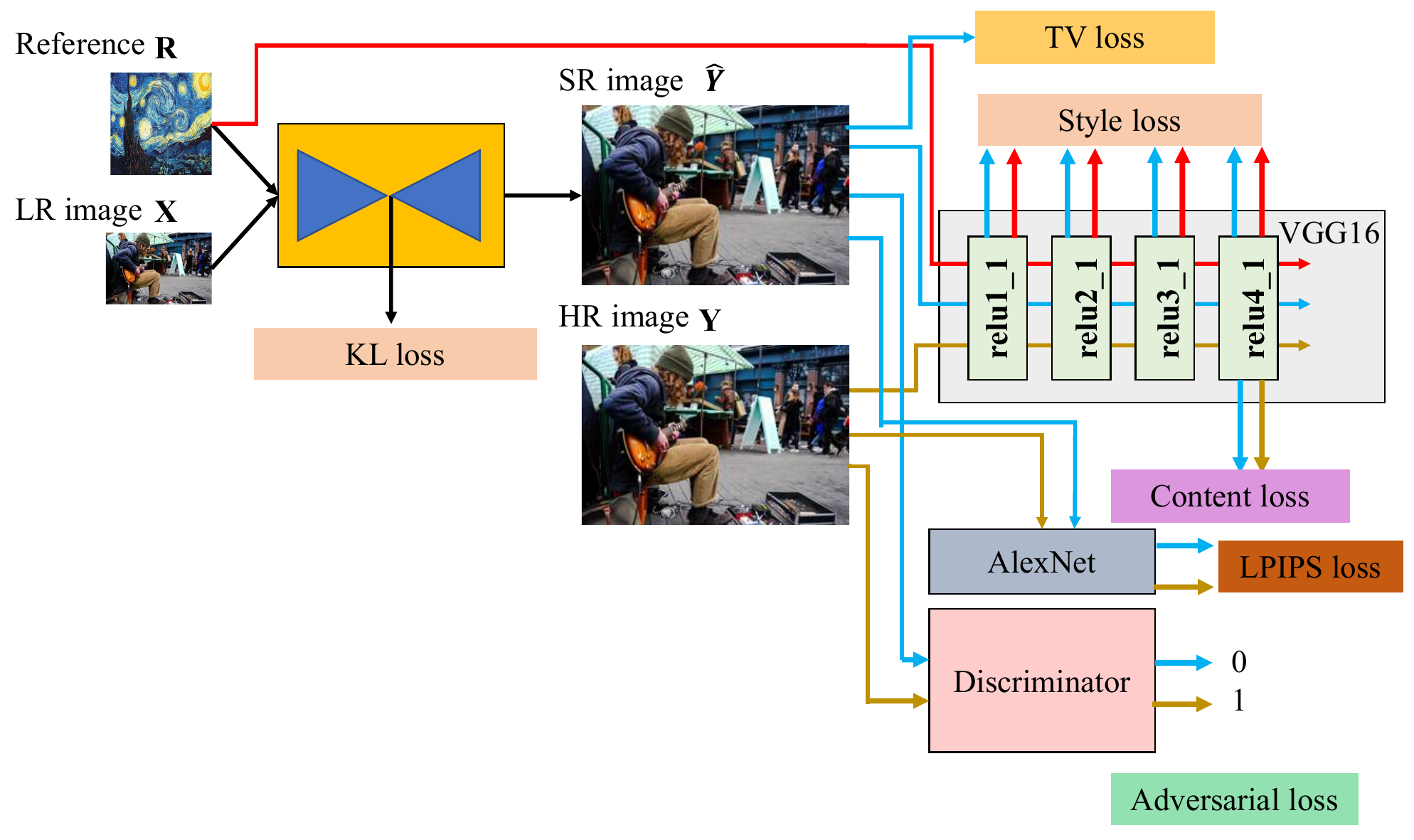}
\end{center}
  \caption{\small{\textbf{Training Losses in the proposed \nnet}. It consists of 1) Content loss, 2) Style loss, 3) KL loss, 4) TV loss, 5) LPIPS loss and 6) Adversarial loss.}
  }
\label{fig:loss}
\vskip -0.1in
\end{figure}

\noindent \textbf{Total loss}. We have the total loss as,

\begin{small}
\vskip -0.2in
\begin{equation} \tag*{(9)}
	\begin{matrix}
	\begin{split}
	L_{\text{style}} = &\lambda_{content} L_{\text{content}} + \lambda_{style} L_{\text{style}} + \lambda_{LPIPS} L_{\text{LPIPS}} \\
	& + \lambda_{TV} L_{\text{TV}} + \lambda_{KL} L_{\text{KL}}~~.  
	\label{eq:Equation9}
	\end{split}
	\end{matrix} 
	\end{equation}
	\vskip -0.2in
\end{small}

\noindent where $\lambda_{content}$, $\lambda_{style}$, $\lambda_{LPIPS}$, $\lambda_{TV}$ and $\lambda_{KL}$ are the weighting parameters for content loss, style loss, LPIPS loss, TV loss and KL loss. Figure~\ref{fig:loss} shows a summary of all the loss terms for readers' reference.

\section{Experiments}
\label{sec:experiments}
\pdfoutput=1
\begin{figure*}[t]
\setlength{\abovecaptionskip}{0pt} 
\begin{center}
\includegraphics[width=\textwidth]{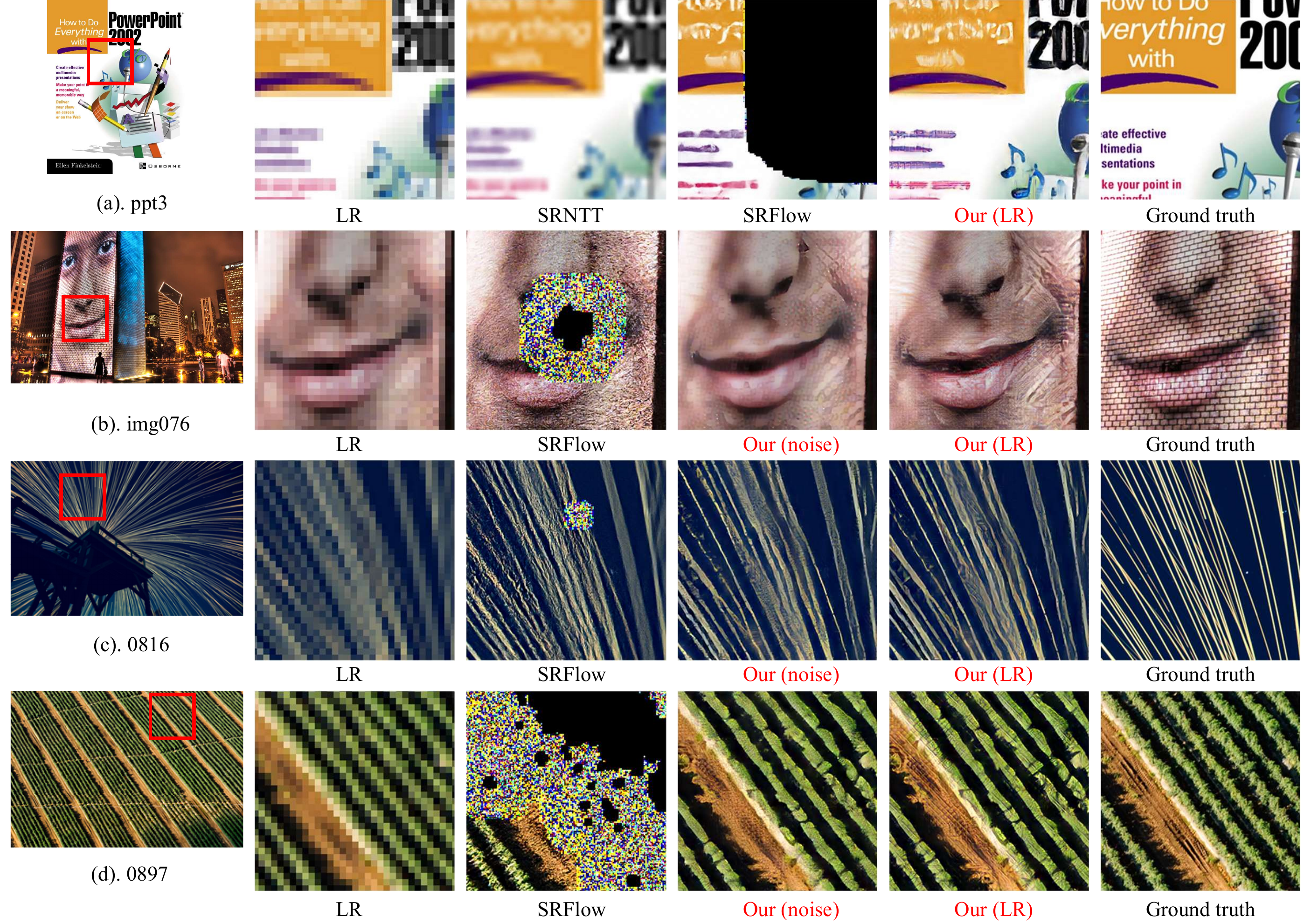}
\end{center}
  \caption{\small{\textbf{Visualization of 8$\times$ SR on different datasets using different SR methods}. (a) is from Set14, (b) is from Urban100, (c) and (d) are from DIV2K validation.}
  }
\label{fig:sota}
\end{figure*} 

\begin{figure*}[t]
\setlength{\abovecaptionskip}{0pt} 
\begin{center}
\includegraphics[width=\textwidth]{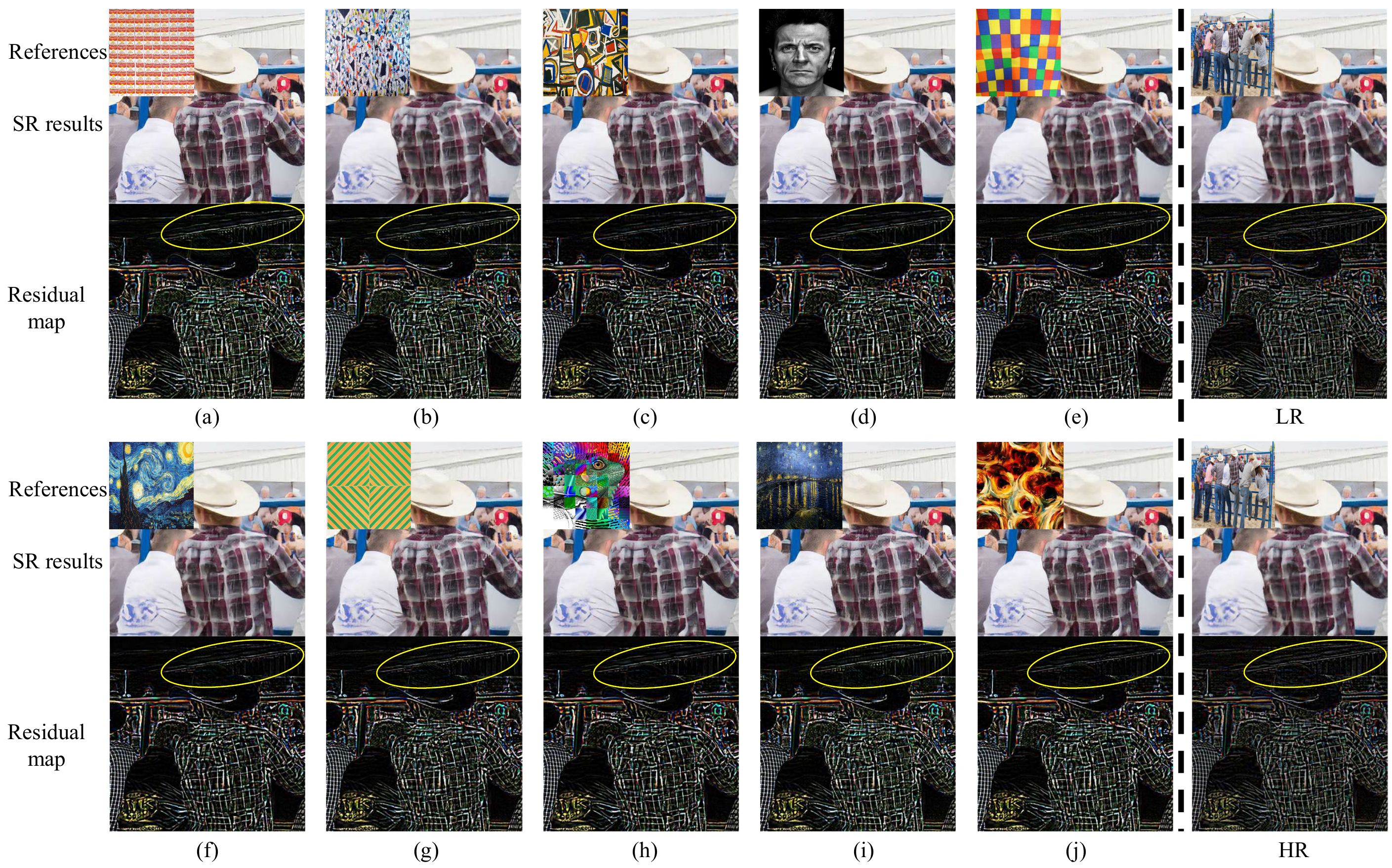}
\end{center}
  \caption{\small{\textbf{Visualization of 8$\times$ SR on using different reference images}. To visualize the differences, we subtract the SR to the ground truth HR image to obtain the residual map. Then we multiple the residual map by 5 to enhance the differences. We also show two examples of using LR and HR images as references for SR. The significant differences are marked in yellow circles.}
  }
\label{fig:multiple}
\vskip -0.01in
\end{figure*} 

\subsection{Data Preparation and Network Details}
We trained our model on DIV2K~\cite{DIV2K} and Flickr2K~\cite{EDSR} datasets. They both contain images with resolution larger than 1000$\times$1000. We extracted LR and HR patches from the training dataset with the size of 32$\times$32 and 32$\alpha\times$32$\alpha$, respectively, where $\alpha$=8 is the upsampling factor. The down-sampling process is done by bicubic interpolation. For the reference image, we used Wikiart~\cite{WikiArt} that is widely used in style transfer. The testing datasets included Set5~\cite{Set5}, Set14~\cite{Set14}, Urban100~\cite{Urban100} and DIV2K validation~\cite{DIV2K}. 

We conducted our experiments using Pytorch 1.7 on a PC with one NVIDIA GTX1080Ti GPU. During the training, we set the learning rate to 0.0001 for all layers. The batch size was set to 16 and we trained it for 5$\times10^4$ iterations. For optimization, we used Adam with the momentum equal to 0.9 and the weight decay of 0.0001. During the training, we set $\lambda_{content}=1$, $\lambda_{style}=10$, $\lambda_{LPIPS}=1$, $\lambda_{TV}=1$ and $\lambda_{KL}=1$, which are the weighting parameters for content loss, style loss, LPIPS loss, TV loss and KL loss. The executive codes and more experimental results can be found in: \url{https://github.com/Holmes-Alan/RefVAE}

\begin{table*}[t]
	\caption{\small{\textbf{Comparison with state-of-the-art methods}. We compare with SRNTT and SRFlow on 8$\times$ super-resolution on PSNR (dB), SSIM and LPIPS.}}
	\label{tab:sota}
	\vspace{-4mm}
	\begin{center}
		\begin{small}
            \resizebox{\linewidth}{!}{			
\begin{tabular}{c|ccc|ccc|ccc|ccc|ccc}
\hline \hline
\multirow{2}{*}{Methods} & \multicolumn{3}{c|}{Set5} & \multicolumn{3}{c|}{Set14} & \multicolumn{3}{c|}{BSD100} & \multicolumn{3}{c|}{Urban100} & \multicolumn{3}{c}{DIV2K validation} \\ \cline{2-16} 
                         & \textit{PSNR} $\uparrow$    & \textit{SSIM} $\uparrow$  & LPIPS $\downarrow$ & \textit{PSNR} $\uparrow$   & \textit{SSIM} $\uparrow$   & \textit{LPIPS} $\downarrow$ & \textit{PSNR} $\uparrow$   & \textit{SSIM} $\uparrow$   & \textit{LPIPS} $\downarrow$  & \textit{PSNR} $\uparrow$     & \textit{SSIM} $\uparrow$   & \textit{LPIPS} $\downarrow$  & \textit{PSNR} $\uparrow$      & \textit{SSIM} $\uparrow$      & \textit{LPIPS} $\downarrow$     \\ \hline
Bicubic                  & 24.39   & 0.657  & 0.537  & 23.19   & 0.568   & 0.630  & 23.67   & 0.547   & 0.713   & 21.24     & 0.516   & 0.686   & 25.17      & 0.664      & 0.584      \\ \hline
SRNTT(HR)                & 24.69   & 0.674  & 0.492  & 23.16   & 0.564   & 0.613  & 23.63   & 0.542   & 0.687   & -         & -       & -       & -          & -          & -          \\ 
SRNTT(LR)                & 24.36   & 0.655  & 0.522  & 23.16   & 0.564   & 0.613  & 23.65   & 0.543   & 0.691   & -         & -       & -       & -          & -          & -          \\ \hline
SRFlow                   & 24.18   & 0.650  & 0.237  & 22.13   & 0.513   & 0.331  & 22.96   & 0.499   & 0.425   & 20.676    & 0.531   & 0.300   & 24.53      & 0.616      & 0.272      \\ \hline
\cellcolor{mistyrose}{Our(LR)}                  & \cellcolor{mistyrose}{25.87}   & \cellcolor{mistyrose}{0.723}  & \cellcolor{mistyrose}{\textbf{0.214}}  & \cellcolor{mistyrose}{24.15}   & \cellcolor{mistyrose}{0.598}   & \cellcolor{mistyrose}{\textbf{0.323}}  & \cellcolor{mistyrose}{24.15}   & \cellcolor{mistyrose}{0.558}   & \cellcolor{mistyrose}{\textbf{0.376}}   & \cellcolor{mistyrose}{21.74}     & \cellcolor{mistyrose}{0.575}   & \cellcolor{mistyrose}{0.334}   & \cellcolor{mistyrose}{24.75}      & \cellcolor{mistyrose}{0.641}      & \cellcolor{mistyrose}{0.232}      \\
\cellcolor{mistyrose}{Our(HR)}                  & \cellcolor{mistyrose}{25.92}   & \cellcolor{mistyrose}{\textbf{0.724}}  & \cellcolor{mistyrose}{0.218}  & \cellcolor{mistyrose}{24.20}   & \cellcolor{mistyrose}{\textbf{0.601}}   & \cellcolor{mistyrose}{0.327}  & \cellcolor{mistyrose}{24.18}   & \cellcolor{mistyrose}{\textbf{0.561}}   & \cellcolor{mistyrose}{0.386}   & \cellcolor{mistyrose}{21.80}     & \cellcolor{mistyrose}{0.575}   & \cellcolor{mistyrose}{\textbf{0.332}}   & \cellcolor{mistyrose}{24.94}      & \cellcolor{mistyrose}{\textbf{0.650}}      & \cellcolor{mistyrose}{\textbf{0.201}}      \\
\cellcolor{mistyrose}{Our(random)}              & \cellcolor{mistyrose}{\textbf{25.94}}   & \cellcolor{mistyrose}{0.721}  & \cellcolor{mistyrose}{0.224}  & \cellcolor{mistyrose}{\textbf{24.28}}   & \cellcolor{mistyrose}{\textbf{0.601}}   & \cellcolor{mistyrose}{0.328}  & \cellcolor{mistyrose}{\textbf{24.25}}   & \cellcolor{mistyrose}{\textbf{0.561}}   & \cellcolor{mistyrose}{0.384}   & \cellcolor{mistyrose}{\textbf{21.85}}     & \cellcolor{mistyrose}{\textbf{0.576}}   & \cellcolor{mistyrose}{0.358}   & \cellcolor{mistyrose}{\textbf{25.00}}      & \cellcolor{mistyrose}{\textbf{0.650}}      & \cellcolor{mistyrose}{0.202}      \\ \hline \hline
\end{tabular}
			}
		\end{small}
	\end{center}
	\vskip -0.1in
\end{table*}

\paragraph{Metrics.}
For evaluation, we have used several metrics: 
\textbf{PSNR} measures the average pixel differences between ground truth and estimation. 
\textbf{SSIM}~\cite{ssim} measures the structural similarity between ground truth and estimation. \footnote{To calculate PSNR and SSIM, we first convert the SR and ground truth images from RGB to YCbCr and take the Y channel for computation.}
\textbf{LPIPS}~\cite{LPIPS} measures the perceptual similarity between ground truth and estimation.
\textbf{Diverse score} measures the spanning of the SR space. We follow the same measurement defined by NTIRE2021 SR space challenge~\cite{ntire2021}. We sampled 10 images, and densely calculated a metric between the samples and the ground truth. To obtain the local best we pixel-wisely selected the best score out of the 10 samples and took the full image's average. The global best was obtained by averaging the whole image's score and selecting the best. Finally, we calculated the score as: (global best - local best)/(global best) $\times$ 100.

\subsection{Compare with state-of-the-art methods}
Our proposed \nnet is for diverse image super-resolution that is able to generate many SR results. We consider two types of related SR methods for comparison: reference based image SR and generative model based image SR. The SR approaches we compare with are:
\textbf{SRNTT}~\cite{refsr4}, reference based SR using similar image as reference for online warping,
and \textbf{SRFlow}~\cite{SRFlow}, normalizing flow based SR that predicts multiple SR results. \footnote{Note that since most of state-of-the-art SR approaches are dedicated for producing one prediction with high PSNR and do not consider the possibility of multiple SR results, it is not our intention to compete with those approaches in terms of PSNR and SSIM.}

In Table~\ref{tab:sota}, we show the overall comparison with SRNTT and SRFlow on 8$\times$ image super-resolution on Set5, Set14, BSD100, Urban100 and DIV2K validation datasets. It is seen that our proposed \nnet can achieve the best PSNR, SSIM and LPIPS by about 2 dB, 0.1 and 0.01, respectively. Note that SRNTT~\cite{refsr4} was originally trained and tested on specific CUFED5 dataset~\cite{refsr4} for which each image has five images with similar contents. For a fair comparison, we tested SRNTT by using LR and HR images as reference for SR, which are SRNTT(LR) and SRNTT(HR). \footnote{Note also that SRNTT requires a lot of memory for online warping. Larger image SR cannot be done on one GPU, hence we did not have the results for Urban100 and DIV2K.} Similarly, we also tested our proposed \nnet by using LR and HR images as Our(LR) and Our(HR), respectively. Visually, we show four examples in Figure~\ref{fig:sota}. It can be seen that our proposed method can restore the details better than others. Especially, it is seen that SRFlow collapses on several cases. For example, SRFlow generated a large black region on \textit{ppt3}. It also produces holes on \textit{img076}, \textit{0816} and \textit{0897} with noises. For SRNTT, it generated blurry results without restoring sharp edges. On the other hand, our proposed \nnet can generate plausible SR results with sharp edges and textures.

More importantly, we are interested in the ability of expansion of SR space. Since our proposed \nnet can take arbitrary references for super-resolution, we can measure the diversity by using \textbf{Diverse score} defined in NTIRE2021 SR space challenge~\cite{ntire2021}. 

\begin{table}[t]
	\caption{\small{\textbf{Analysis on SR space} of \nnet on DIV2K dataset. The PSNR, SSIM and LPIPS on SR estimation are indicated as \textit{SR PSNR}, \textit{SR SSIM} and \textit{SR LPIPS}, while for LR estimation they are respectively indicated as \textit{LR PSNR} and \textit{LR SSIM}. The overall estimation using SRFlow and our methods are shown in the 2nd half of the table.}}
	\label{tab:div}
	\vspace{-4mm}
	\begin{center}
		\begin{small}
\resizebox{\linewidth}{!}{			
\begin{tabular}{cccccc}
\hline \hline
\multicolumn{1}{c|}{Reference}                                                                 & \textit{SR PSNR} $\uparrow$ & \textit{SR SSIM} $\uparrow$ & \textit{SR LPIPS} $\downarrow$ & \textit{LR PSNR} $\uparrow$ & \textit{LR SSIM} $\uparrow$ \\ \hline
\multicolumn{1}{c|}{Ref. 1}                                                                & 24.27   & 0.616   & 0.310    & 42.90   & 0.996   \\
\multicolumn{1}{c|}{Ref. 2}                                                                & 25.18   & 0.661   & 0.306    & 44.55   & 0.997   \\
\multicolumn{1}{c|}{Ref. 3}                                                                & 25.03   & 0.654   & 0.304    & 44.37   & 0.997   \\
\multicolumn{1}{c|}{Ref. 4}                                                                & 24.76   & 0.644   & 0.302    & 43.93   & 0.997   \\
\multicolumn{1}{c|}{Ref. 5}                                                                & 25.28   & 0.664   & 0.314    & 44.80   & 0.997   \\
\multicolumn{1}{c|}{Ref. 6}                                                                & 25.17   & 0.662   & 0.306    & 44.81   & 0.997   \\
\multicolumn{1}{c|}{Ref. 7}                                                                & 24.62   & 0.627   & 0.298    & 43.43   & 0.996   \\
\multicolumn{1}{c|}{Ref. 8}                                                                & 25.04   & 0.659   & 0.307    & 44.26   & 0.997   \\
\multicolumn{1}{c|}{Ref. 9}                                                                & 24.27   & 0.611   & 0.320    & 42.86   & 0.996   \\
\multicolumn{1}{c|}{Ref. 10}                                                               & 24.42   & 0.615   & 0.310    & 42.30   & 0.996   \\
\multicolumn{1}{c|}{LR}                                                               & 25.31   & 0.664   & 0.306    & 46.21   & 0.997   \\
\multicolumn{1}{c|}{HR}                                                               & 25.40   & 0.665   & 0.308    & 45.63   & 0.997   \\ \hline \hline
\multicolumn{6}{c}{Overall Estimation}                                                                                                        \\ \hline \hline
\multicolumn{1}{c|}{\begin{tabular}[c]{@{}c@{}}SRFlow\\ Div. score: 10.07 \end{tabular}}          &  24.46  & 0.615 & 0.328  & \textbf{51.18}   & 0.996 \\ \hline
\multicolumn{1}{c|}{\begin{tabular}[c]{@{}c@{}}Our (Reference)\\ Div. score: 14.80 \end{tabular}} & 24.89   & 0.641   & \textbf{0.308}    & 44.17   & \textbf{0.997}   \\ \hline
\multicolumn{1}{c|}{\begin{tabular}[c]{@{}c@{}}Our (Random)\\ Div. score: \textbf{14.91} \end{tabular}}    & \textbf{24.91}   & \textbf{0.642}   & 0.321    & 44.77   & \textbf{0.997}   \\ \hline \hline
\end{tabular}
			}
		\end{small}
	\end{center}
	\vskip -0.2in
\end{table}

As discussed in Section III, our proposed \nnet can take random noise or external images as references. In Table~\ref{tab:div}, we randomly chose 10 images from WikiArt as references for SR. We also use LR image itself and HR image as references. We list the results on SR estimation as \textit{SR PSNR}, \textit{SR SSIM} and \textit{SR LPIPS}. It can be seen that the proposed \nnet can generate different SR images (different PSNR and SSIM) given different references. It demonstrates that the proposed \nnet can expand the SR space by a large margin. In the meantime, we do not want to distort SR images to be far away from the ground truth LR images. Hence, we down-sample SR images by bicubic to estimate PSNR and SSIM on the LR space as \textit{LR PSNR} and \textit{LR SSIM}, respectively. It can be seen that different SR estimation can well preserve LR information with PSNR over 40 dB. 

Overall, we have calculated the \textbf{Diverse score} of SRFlow and our approach in Table~\ref{tab:div}. Given different references, we have two results as entitled Our(Reference) and Our(Random). It can be seen that our approach can outperform SRFlow in terms of PSNR, SSIM and \textbf{Diverse score}. We also show SR results of using 12 different references in Figure~\ref{fig:multiple}. To visualize the differences among multiple SR images, we subtracted the SR by the ground truth HR images to obtain the residual maps, then we multiplied the residual values by 5 to show the differences. We can observe the differences around the edge and texture regions from the residual maps (marked in yellow circles), e.g., the patterns of the T-shirt and the roof of the building. In the meantime, we also visualize the SR results using LR and HR images. It can be seen that using LR or HR as references can better reduce the residues.

\begin{table}[t]
	\caption{\small{\textbf{Ablation study} of \nnet on DIV2K dataset on 8$\times$ super-resolution. We report the average across datasets.}}
	\label{tab:ablation}
	\vspace{-4mm}
	\begin{center}
		\begin{small}
\resizebox{\linewidth}{!}{			
\begin{tabular}{c|ccc|cccc}
\hline \hline
\multirow{2}{*}{Model} & \multicolumn{3}{c}{Components} & \multicolumn{4}{c}{Eval.} \\ \cline{2-8} 
                       & CVAE     & SC     & Dicri.     & \textit{PSNR} $\uparrow$   & \textit{SSIM} $\uparrow$   & \textit{LPIPS} $\downarrow$  & \textit{Div.} $\uparrow$ \\ \cline{2-8} 
\multirow{3}{*}{Our}   & -        & -      & -          & \textbf{25.23}   & \textbf{0.650}   & 0.412  & 0\\
                       &  \checkmark  & -      & -          & 24.82   & 0.641   & 0.338  & 12.326 \\
                       &  \checkmark   &  \checkmark  & -          & 24.84   & 0.643   & 0.400  & 14.142\\ \hline
\cellcolor{mistyrose}{Final}                  &  \cellcolor{mistyrose}{\checkmark}   &  \cellcolor{mistyrose}{\checkmark}  & \cellcolor{mistyrose}{\checkmark}      & \cellcolor{mistyrose}{24.75}   & \cellcolor{mistyrose}{0.640}   & \cellcolor{mistyrose}{\textbf{0.272}}  & \cellcolor{mistyrose}{\textbf{14.992}}\\ \hline \hline
\end{tabular}
			}
		\end{small}
	\end{center}
	\vskip -0.2in
\end{table}

\subsection{Ablation study}
Let us evaluate various key components of the \nnet:  
(1) the Conditional Variational AutoEncoder (CVAE),  
(2) the use of style and content losses (SC loss), 
(3) the use of discriminator (Discri.), 
To examine their impact, we started with a simple network structure consisting Image Encoder and Decoder only, and we progressively added all components. Table~\ref{tab:ablation} reports the results of 8$\times$ SR on DIV2K validation datasets. We have included PSNR, SSIM, LPIPS and \textit{Div.} (\textbf{Diverse score}) in the table. From the values of \textit{Div.}, we can find that using CVAE module can introduce randomness into the network to generate multiple SR results. When the style and content losses were used for training, we can further maximize the \textbf{Diverse score} to expand the SR space. From the values of PSNR, SSIM and LPIPS, using CVAE, SC loss and Discri can reduce the PSNR and SSIM, but they can improve the LPIPS value, which indicates better quality.

\begin{figure}[t]
\begin{center}
\includegraphics[width=1\columnwidth]{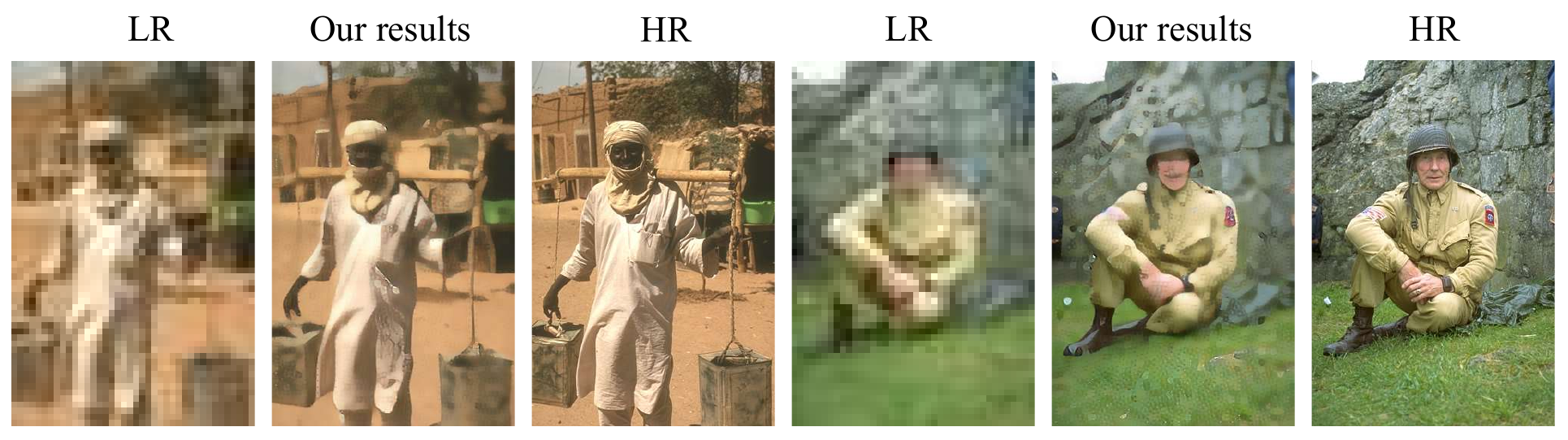}
\end{center}
  \caption{\small{\textbf{Weak cases of using proposed \nnet}. We show two 8$\times$ SR examples from BSD100 dataset. It can be found that the facial details cannot be preserved.}
  }
\label{fig:failure}
\end{figure} 

\subsection{Weak cases}

In terms of perceptual quality, our proposed method does not perform too well on images with smaller size. We show two examples in Figure~\ref{fig:failure}. It can be seen that the fine facial expression cannot be ideally restored by our approach. To resolve this problem, we could further study in depth of the latent representation of the Variational AUtoEncoder. One other solution is to use hierarchical latent space manipulation. This is really a fruitful direction for research.

\section{Conclusion}
\label{sec:conclusion}
\pdfoutput=1
In this paper, we introduce a novel approach for reference based image super-resolution \nnet. Unlike other reference based SR approaches, our proposed method can take any image with arbitrary content or even random noise as reference for image super-resolution. It is also unlike other generative model based SR approaches, \nnet can expand the SR space, so that multiple unique SR images can be generated. Furthermore, \nnet does not suffer from the model collapse or generate bizarre patterns like other generative model based approaches (examples are shown in Figure~\ref{fig:sota}). Our analysis shows that \nnet can better preserve the LR information and generate SR image with better perceptual quality. Possible future work includes the super-resolution space exploration by probing the latent vector or by image quantization.

\newpage
{\small
\bibliographystyle{ieee_fullname}
\bibliography{egbib}
}

\end{document}